\documentclass[journal,twoside,web]{ieeecolor}

\usepackage{tmi}
\usepackage{cite}
\usepackage{amsmath,amssymb,amsfonts}
\usepackage{algorithmic}
\usepackage[linesnumbered, ruled]{algorithm2e}
\usepackage{graphicx}
\usepackage{textcomp}
\usepackage[colorlinks,linkcolor=red]{hyperref}

\usepackage{multirow}
\usepackage{booktabs}
\usepackage{bbm}

\def\BibTeX{{\rm B\kern-.05em{\sc i\kern-.025em b}\kern-.08em
    T\kern-.1667em\lower.7ex\hbox{E}\kern-.125emX}}
\begin{document}
\title{A simple normalization technique using window statistics to improve the out-of-distribution generalization on medical images}
\author{Chengfeng Zhou, Songchang Chen, Chenming Xu, Jun Wang, Feng Liu, Chun Zhang, Juan Ye, Hefeng Huang, and Dahong Qian, Senior Member, IEEE

\thanks{This work was supported by the National Natural Science
Foundation of China (Grant No. 81974276), the Science and Technology Commission of Shanghai Municipality under Grant 20DZ2220400, and the Deepwise Joint Lab. (Corresponding authors: J. Ye, H. Huang, and D. Qian.)}
\thanks{C. Zhou and D. Qian are with the School of Biomedical
Engineering, Shanghai Jiao Tong University, Shanghai, China (e-mail: joe1chief1993@gmail.com; dahong.qian@sjtu.edu.cn).}
\thanks{S. Chen, C. Xu, and H. Huang are with the Obstetrics and Gynecology Hospital, Institute of Reproduction and Development, Fudan University, Shanghai,
China (e-mail: 26775172@qq.com; xuchenm@163.com; huanghefg@fudan.edu.cn)}
\thanks{J. Wang is with the Zhejiang University City College, Hangzhou, China (e-mail: wjcy19870122@163.com).}
\thanks{F. Liu is with the Deepwise AI Lab, Beijing, China (e-mail: liufeng@deepwise.com).}
\thanks{J. Ye is with the Department of Ophthalmology, the Second Affiliated Hospital of Zhejiang University, Hangzhou, China (e-mail: yejuan@zju.edu.cn).}
\thanks{C. Zhang is with the Department of Ophthalmology, Peking University Third Hospital, Beijing, China (e-mail: zhangc1@yahoo.com).}
}

\maketitle

\begin{abstract}

Since data scarcity and data heterogeneity are prevailing for medical images, well-trained Convolutional Neural Networks (CNNs) using previous normalization methods may perform poorly when deployed to a new site. However, a reliable model for real-world clinical applications should be able to generalize well both on in-distribution (IND) and out-of-distribution (OOD) data (e.g., the new site data). In this study, we present a novel normalization technique called window normalization (\emph{WIN}) to improve the model generalization on heterogeneous medical images, which is a simple yet effective alternative to existing normalization methods. Specifically, \emph{WIN} perturbs the normalizing statistics with the local statistics computed on the window of features. This feature-level augmentation technique regularizes the models well and improves their OOD generalization significantly. Taking its advantage, we propose a novel self-distillation method called \emph{WIN-WIN} for classification tasks. \emph{WIN-WIN} is easily implemented with twice forward passes and a consistency constraint, which can be a simple extension for existing methods. Extensive experimental results on various tasks (6 tasks) and datasets (24 datasets) demonstrate the generality and effectiveness of our methods.

\end{abstract}


\section{Introduction}
\label{sec:introduction}
\begin{figure}[tb]
    \begin{center}
    \includegraphics[width=\linewidth]{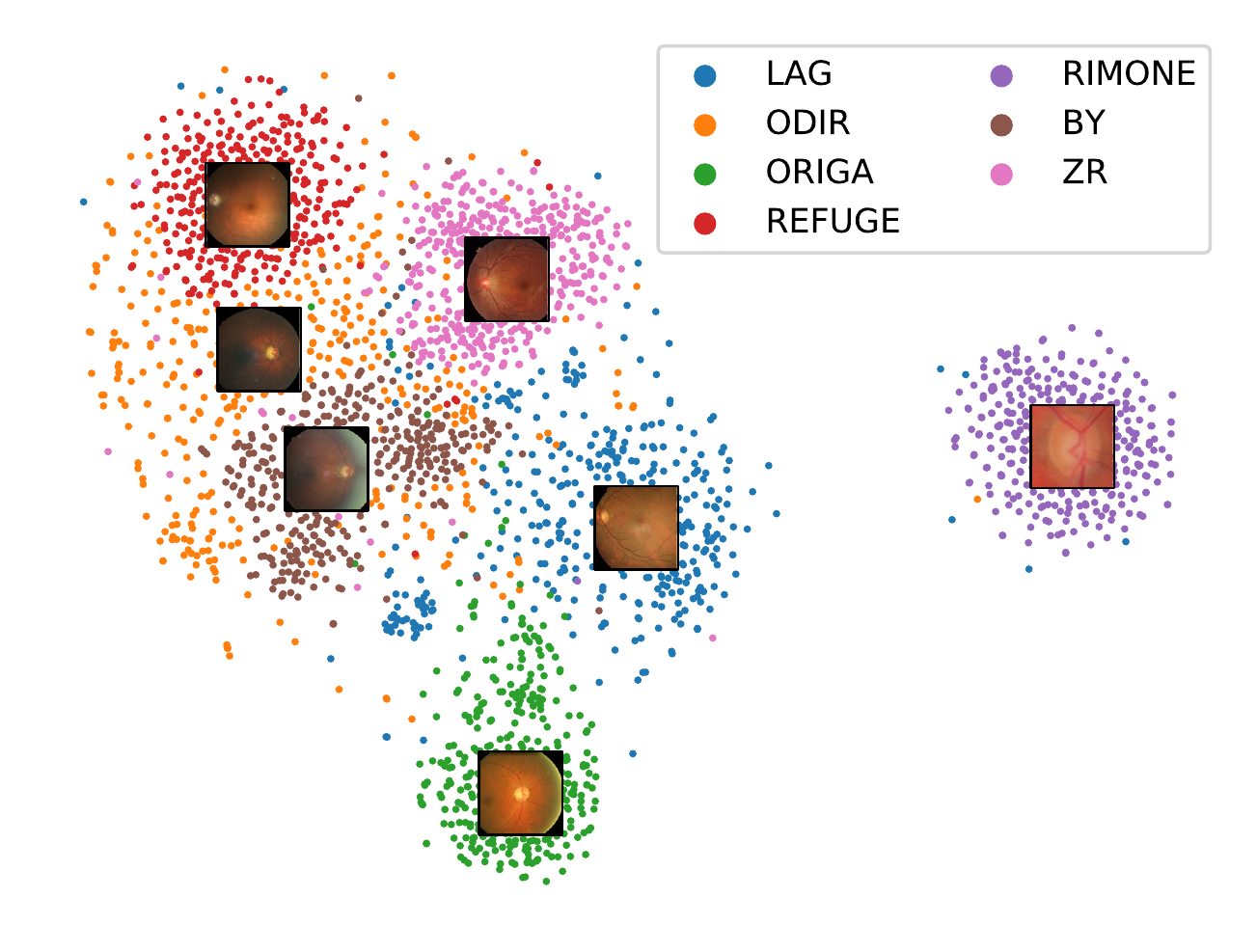}
    \vspace{-0.9cm}
    
    \caption{ T-SNE visualization of features on seven glaucoma detection datasets. Each dataset only captures a small population of real data. Thus, the distribution shift for images exists between any two datasets.} 
    \label{fig:domain_gap} 
    \end{center}
    \vspace{-1.0cm}
    
\end{figure}

\IEEEPARstart{D}{e}SPITE the tremendous success of CNNs in medical image analysis, they are mainly built upon the “i.i.d. assumption”,  which states that training data and test data are independent and identically distributed. The assumption barely holds in real-world applications due to the nature of medical images, rendering the sharp drop of well-trained models on unseen data with distribution shifts \cite{hendrycks2021many, koh2021wilds, taori2020measuring}. Since the expensive costs for data acquisition and data annotation lead to data scarcity, the data used for model training can only capture a small population of real data distribution (see Fig. \ref{fig:domain_gap}). Meanwhile, the data heterogeneity caused by inconsistent standards (e.g., various operating procedures and imaging equipment) worsens the distribution shifts in medical images. A slight difference in appearance results in significant model performance dips \cite{koh2021wilds}. Therefore, OOD generalization on heterogeneous data is a crucial challenge for real-world clinical applications \cite{ShujunWang2020DoFEDF, li2021fedbn}.

The most straightforward solution is to acquire sufficient data from various sites and train a robust model. However, it would be high-cost and even impossible in real-world applications. In practice, another economic and popular solution is data augmentation. Data augmentation enhances the breadth of seen data with predefined class-preserving options, such as mixing multiple augmented images \cite{DBLP:conf/iclr/HendrycksMCZGL20} and mixing an augmented image with dreamlike images \cite{hendrycks2022robustness}. With the help of the augmented samples simulating the unseen images, CNNs improve their generalization to OOD data. Although the data augmentation methods no longer require significant manual  designation efforts \cite{cubuk2019autoaugment, cubuk2020randaugment}, their performance is less appealing for improving the OOD generalization on medical images because they are mainly developed for the natural images with large domain gaps \cite{zhou2021mixstyle, KaiyangZhou2020LearningTG, Li2021SFA}. Besides, they may increase computing overhead and impede model convergence. 

Normalization layers are essential components of modern CNNs. Traditional normalization techniques are generally and simplistically built upon the “i.i.d. assumption”. For instance, Batch Normalization (\emph{BN}) \cite{DBLP:conf/icml/IoffeS15} constrains intermediate features within the normalized distribution with mini-batch statistical information to stabilize and accelerate the training. It has a fundamental flaw, train-test statistics inconsistency, which worsens under distribution shifts. Instance normalization (\emph{IN}) \cite{DBLP:journals/corr/UlyanovVL16} overcomes this limitation through computing the normalization statistics consistently. During the training and testing, \emph{IN} normalizes features with statistics of the spatial dimension. This procedure effectively reduces the data distribution (i.e., by eliminating the style discrepancy) and improves the OOD generalization \cite{DBLP:conf/eccv/PanLST18}. But its improvement for OOD generalization is still inadequate. On this basis, subsequent studies directly combine it with other mechanisms to address the OOD generalization problem, but they are complicated and inefficient for medical images \cite{DBLP:conf/eccv/PanLST18, tang2021crossnorm, jin2021style, luo2019switchable, nam2018batch, WenqiShao2019SSNLS}. For instance, Jin et al. \cite{jin2021style} designed a normalization technique that consists of three components: Style Normalization and Restitution Module, Dual Causality loss, and Dual Restitution Loss.

In this work, we implement the feature-level data augmentation by a simple yet effective normalization technique called Window Normalization (\emph{WIN}) to improve the OOD generalization on heterogeneous medical images. \emph{WIN} exploits the mean and variance of a stochastic window to perturb the normalization statistics with a mixup operation, which is extremely easy to use and remarkably effective. With the benefits of \emph{WIN}, we introduce a novel self-distillation scheme \emph{WIN-WIN} which forward passes the input in different model modes twice and compels the consistency between the outputs. \emph{WIN-WIN} can be implemented with few codes and further improve the OOD generalization in classification tasks. We demonstrate that our methods can generally boost the OOD generalization of various tasks (such as glaucoma detection, breast cancer detection, chromosome classification, optic disc and cup segmentation, etc.), spanning $24$ datasets, with free parameters and a negligible effect for IND data. Our main contributions are summarized as follows:
\begin{itemize}
    \item  We develop a simple normalization \emph{WIN} for improving OOD generalization on heterogeneous data. \emph{WIN} is a good alternative for existing normalization methods, with free parameters and a negligible effect for IND data.
    \item We propose a novel self-distillation scheme \emph{WIN-WIN} to fully exploit the \emph{WIN} in classification tasks, which is easy to implement and significantly effective.
    \item  We demonstrate through extensive experiments that our methods can significantly and generally boost the OOD generalization across various tasks. The code of our method is publicly available at \url{https://github.com/joe1chief/windowNormalizaion}.
\end{itemize}

\section{Related work}
\label{sec:relatedWork}

\subsection{Data Augmentation}

Data augmentation is an important tool in training deep models, which effectively increases the data amount and enriches the data diversity through random operations such as translation, flipping, and cropping \cite{krizhevsky2012imagenet}. Intuitively, simulated and augmented inputs or outputs could aid a model in learning the invariances among the data domains and improve its generalization to novel domains. Popular data augmentation techniques are conducted at either image-level or feature-level. Image-level augmentation usually needs manual designation, which is challenging due to the difficulty of simulating the data from unforeseen target domains. To this end, several methods automatically searched for the augmentation policies in their predefined search space \cite{cubuk2019autoaugment, cubuk2020randaugment}. However, their performance on OOD data is doubtful as the augmentation policies were usually tuned to optimize performance on a particular domain. 

In the OOD generalization literature, Lopes et al. \cite{lopes2019improving} added noise to randomly selected patches on the input image. AugMix \cite{DBLP:conf/iclr/HendrycksMCZGL20} and PIXMIX \cite{hendrycks2022robustness} mixed multiple augmented images or an augmented image and dreamlike images, respectively. Zhou et al. \cite{KaiyangZhou2020LearningTG} synthesized the pseudo-novel domain data with a data generator. DeepAugment \cite{hendrycks2021many} perturbed the input image with an image-to-image network. In contrast, feature-level data augmentation has also worked well in this context. Verma et al. \cite{verma2019manifold} simply mixed up the latent features. Zhou et al. \cite{zhou2021mixstyle} designed a plug-and-play module that mixes feature statistics between instances. Li et al. \cite{Li2021SFA} introduced an augmentation module that perturbs the feature with a data-independent noise and an adaptive dependent noise.

However, the aforementioned methods inevitably introduce computational overhead. Our method implements the feature-level data augmentation by the essential normalization layers in CNNs and outperforms many prior data augmentation techniques for improving OOD generalization.

\subsection{Normalization}
Normalization layer is an essential module in deep models. Extensive normalization techniques have been proposed for various domains such as natural language processing, computer vision, and machine learning.  A milestone study is Batch Normalization (\emph{BN}) \cite{DBLP:conf/icml/IoffeS15}, which has verified its generality on a wide variety of tasks. Although \emph{BN} has been empirically proven to accelerate the model convergence and combat overfitting, it exhibits several shortcomings because it computes the statistic along the batch dimension by default \cite{wu2018group, wu2021rethinking}. For example, the instance-specific information is lacking. To this end, Li et al. \cite{li2020attentive}  re-calibrated features via an attention mechanism and Gao et al. \cite{gao2021representative} added a feature calibration scheme into \emph{BN} to incorporate this information. Instance Normalization (\emph{IN}) \cite{DBLP:journals/corr/UlyanovVL16} is another milestone study that normalizes features with statistics of the spatial dimension. It has been extensively applied in the field of image style transfer as it could eliminate instance-specific style discrepancy (namely, standardizing features with the mean and the variance of \emph{IN}) \cite{huang2017arbitrary}. However, \emph{IN} eliminated the style information and resulted in inferior performance on in-distribution samples \cite{DBLP:conf/eccv/PanLST18, jin2021style}. Following the \emph{BN} and \emph{IN}, Group Normalization (GN) \cite{wu2018group} exploited the statistics of grouped channel dimensions to train the model with small batches. On the contrary, there are several normalization techniques that utilized statistics of partial regions \cite{ortiz2020local, yu2020region} instead of all pixels within a dimension (e.g., Batch Normalization, Instance Normalization, Layer Normalization \cite{ba2016layer}, and Group Normalization). Ortiz et al. \cite{ortiz2020local} proposed the local context normalization (LCN) in which each feature is normalized based on a window around it and the filters in its group. It has demonstrated its effectiveness in dense prediction tasks, including object detection, semantic segmentation, and instance segmentation applications.

Since it has been a subtle choice about which normalization techniques can be applied in a particular scenario, many studies have explored the combination of multiple normalizations. Batch-Instance Normalization (BIN) \cite{nam2018batch} adaptively balanced the \emph{BN} output and the \emph{IN} output with a learnable gate parameter. This gating mechanism also appeared in the Switchable Normalization (SN) series \cite{luo2019switchable, WenqiShao2019SSNLS} which combined three types of statistics estimated channel-wise \cite{DBLP:journals/corr/UlyanovVL16}, layer-wise \cite{ba2016layer}, and minibatch-wise \cite{DBLP:conf/icml/IoffeS15}. Besides, Qiao et al. \cite{qiao2019rethinking} introduced another combination that stacked the \emph{BN} with the GN rather than conducting a weighted summation operation. These multi-normalization combination methods enable better adaptability and easy usage for various deep networks and tasks. However, they usually bring the extra computational cost. 

Above normalization techniques are proposed under the assumption that training and test data follow the same distribution. Meanwhile, numerous normalization studies are proposed to address the distribution shifts between training and test distribution in real-world applications. For example, Mode Normalization \cite{deecke2018mode} assigned the samples to different modes and normalized them with their corresponding statistics. Li et al. \cite{li2021fedbn} used local batch normalization to address non-iid training data. Tang et al. \cite{tang2021crossnorm} enlarged the training distribution with CrossNorm and bridged the gaps between training and test distribution with SelfNorm. Since the \emph{IN} can effectively alleviate the style discrepancy among domains/instances, Pan et al. \cite{DBLP:conf/eccv/PanLST18} mixed \emph{IN} and \emph{BN} in the same layer to narrow the distribution gaps and Jin et al. \cite{jin2021style} proposed a style normalization and restitution module to eliminate the style variation and preserve the discrimination capability. Compared to these methods, \emph{WIN} simply yet effectively addresses the OOD generalization problem. It has zero parameters to tune and no need for cautious choices of plug-in manner. As a variant of \emph{IN}, it can be directly employed as the normalization layers and bring free benefits for OOD generalization.

\section{Methods}
\begin{figure}[tb]
    \begin{center}
    \includegraphics[width=\linewidth]{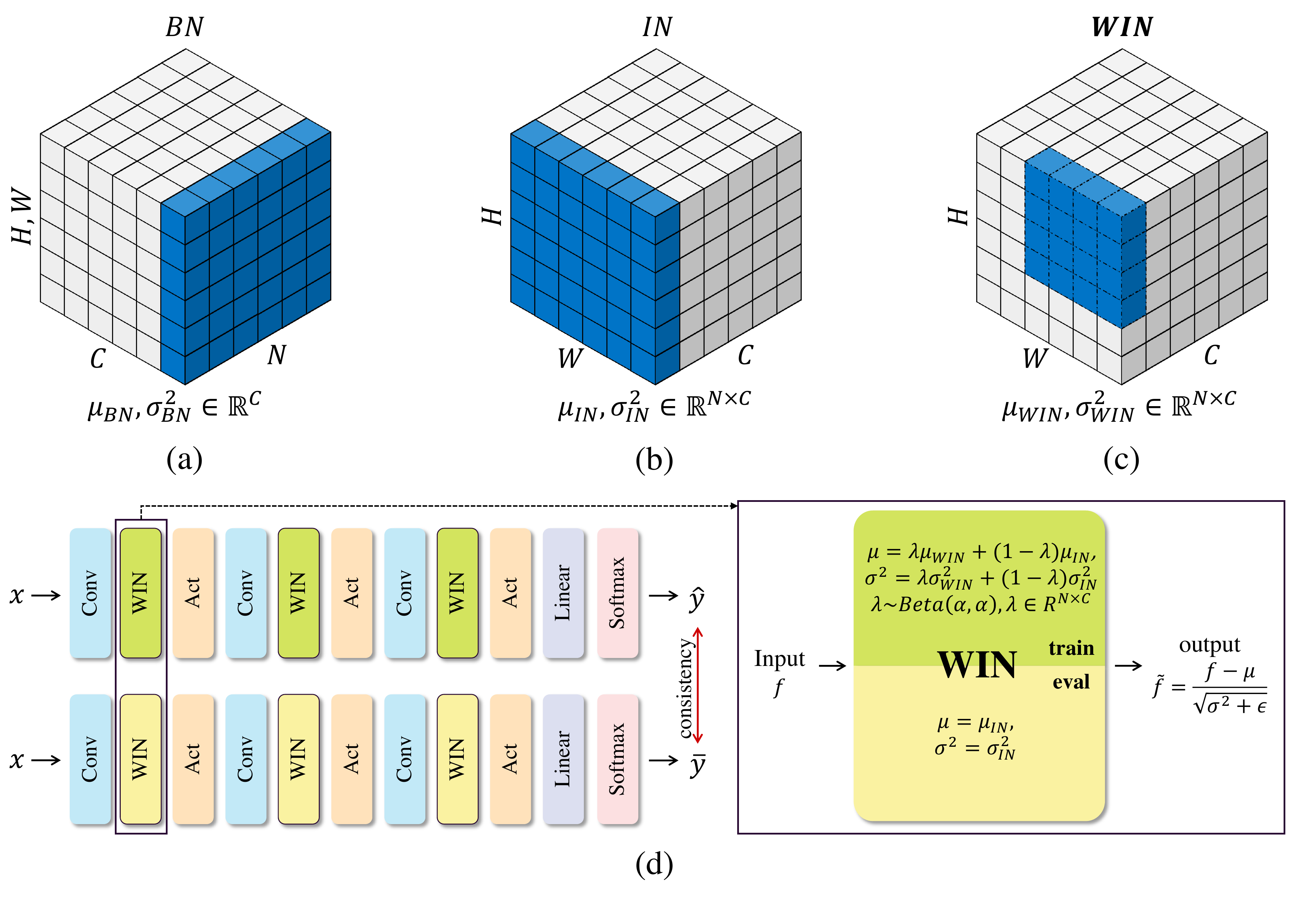}
    \vspace{-1.0cm}
    
    \caption{ (a)-(c) Statistics calculation. Each subplot shows the feature map $f \in \mathbb{R}^{N\times C \times H \times W}$. The mean and variance are computed by aggregating the values of the pixels in blue. And, these pixels share the same normalizing statics. (d) The schematic illustration of \emph{WIN-WIN} (right) and \emph{WIN} (left). \emph{WIN} is adopted as the normalization layer in CNNs. It uses mixing statistics during the training and global statistics during the evaluation. $x$ denotes the inputs. $\hat{y}$ and $\bar{y}$ are the prediction logits in training mode and evaluation mode, respectively. \emph{WIN-WIN} passes the input $x$ twice and compels the consistency between outputs $\hat{y}$ and $\bar{y}$.  } 
    \label{fig:WIN} 
    \end{center}
    \vspace{-0.9cm}
    
\end{figure}

\subsection{Background}
Normalization methods usually consist of feature standardization and affine transformation operations. Given the input feature $f \in \mathbb{R}^{N\times C \times H \times W}$ in 2D CNNs, where $N$, $C$, $H$, and $W$ are batch size, channel number, height, and width of the input feature, respectively. The feature standardization and affine transformation can be formulated as follows:

\begin{align}
    \label{eq:Standardization}
    Standardization : \bar{f} &=\frac{f-\mu}{\sqrt{\sigma^{2}+\epsilon}},\\ 
    Affine : \tilde{f} &= \gamma \odot \bar{f} +\beta.
\end{align}  
$\mu$ and $\sigma^2$ denote the mean and variance, respectively. $\gamma$ and $\beta$ are learnable affine parameters. $\epsilon$ is used to prevent zero variance.

In \emph{BN}, the mean and variance during training are defined as follows: 

\begin{equation}
\begin{aligned}
&\mu_{BN}^{c}=\frac{1}{N H W} \sum_{n=1}^{N} \sum_{h=1}^{H} \sum_{w=1}^{W} f_{n, c, h, w}, \\
&{\sigma_{BN}^{2}}^{c}=\frac{1}{N H W} \sum_{n=1}^{N} \sum_{h=1}^{H} \sum_{w=1}^{W}\left(f_{n, c, h, w}-\mu_{BN}^{c}\right)^{2}.
\end{aligned}
\end{equation}
$\mu_{BN}$, $\sigma_{BN}^{2}$ $\in \mathbb{R}^{C}$ are computed within each channel of $f$ (see Fig. \ref{fig:WIN} (a)). During the evaluation, the mean and variance are computed as exponential moving averages of $\mu_{BN}$, $\sigma_{BN}^{2}$ and accumulated as follows:

\begin{equation}
\begin{aligned}
\mu_{BN}^{\prime} & \Leftarrow p \mu_{BN}^{\prime}+(1-p) \mu_{BN}, \\
{\sigma_{BN}^{2}}^{\prime} & \Leftarrow p {\sigma_{BN}^{2}}^{\prime}+(1-p) \sigma_{BN}^{2},
\end{aligned}
\end{equation}
where $p \in [0,1]$ is a momentum. However, the running mean and variance usually result in significant degradation in performance since they cannot be aligned with the statistics estimated on OOD data \cite{wu2021rethinking}. In addition, for the affine transformation, \emph{BN} employs two learnable parameters $\gamma_{BN} \in \mathbb{R}^{C}$ and $\beta_{BN} \in \mathbb{R}^{C}$.

\emph{IN} defines the mean and variance as follows:
\begin{equation}
\begin{aligned}
&\mu_{IN}^{n,c}=\frac{1}{H W} \sum_{h=1}^{H} \sum_{w=1}^{W} f_{n, c, h, w}, \\
&{\sigma_{IN}^{2}}^{n,c}=\frac{1}{HW} \sum_{h=1}^{H} \sum_{w=1}^{W}\left(f_{n, c, h, w}-\mu_{IN}^{n,c}\right)^{2} .
\end{aligned}
\end{equation}
$\mu_{IN}$, $\sigma_{IN}^{2}$ $\in \mathbb{R}^{N\times C}$ are computed across spatial dimensions independently for each channel and each instance (see Fig. \ref{fig:WIN}(b)). In addition, $\mu_{IN}$ and $\sigma_{IN}^{2}$ encode the instance-specific style information. Thus, standardizing the feature using $\mu_{IN}$ and $\sigma_{IN}^{2}$ is a kind of style normalization. Style normalization will eliminate the feature variance caused by appearance, which benefits the OOD generalization. However, it inevitably removes some discriminative information and degrades the IND generalization \cite{DBLP:conf/eccv/PanLST18}. In the \emph{IN}, the calculation of mean and variance is consistent during the training and evaluation, and the affine transformation which adopts $\gamma_{IN} \in \mathbb{R}^{N \times C}$ and $\beta_{IN} \in \mathbb{R}^{N \times C}$ is usually deactivated in practice.

\begin{algorithm}[tb]
\caption{\emph{Window} sampling}\label{algorithm}
\KwData{Features $f \in \mathbb{R}^{N\times C \times H \times W}$, Threshold $\tau$, \\ Window ratio $\tau^{\prime}$, Window width $\bar{W}$, Window height $\bar{H}$, Window center $(x, y)$;}
\KwResult{Top-left of window $(\bar{x},\bar{y})$, Bottom-right of window $(\tilde{x},\tilde{y})$;}
\Repeat{$(\tilde{x}-\bar{x})\times (\tilde{y}-\bar{y}) \geq \tau \times H \times W$}{
      \tcp{get a squared window}
      Uniformly sample a window ratio $\tau^{\prime} \sim U(0.0, 1.0)$\;
      $\bar{W} \gets W * sqrt(\tau^{\prime})$\; 
      $\bar{H} \gets H * sqrt(\tau^{\prime})$\; 
      Uniformly sample a window center $(x, y)$\;
      \tcp{place the window}
      $\bar{x} \gets int(max(min(x-\bar{W}//2,0), W))$\; 
      $\bar{y} \gets int(max(min(y-\bar{H}//2,0), H))$\;
      $\tilde{x} \gets int(max(min(x+\bar{W}//2,0), W))$\;
      $\tilde{y} \gets int(max(min(y+\bar{H}//2,0), H))$\; 
    }

\end{algorithm}

\subsection{Window Normalization}
Our main inspiration is the observation of Ghost Batch Normalization (GBN) \cite{dimitriou2020new} which improves the model generalization with the mean and variance calculated on small parts of the input batch. Intuitively, GBN is equivalent to adding different noises to different slices of a batch, which is a kind of feature-level data augmentation. On the other hand, \emph{IN} demonstrated its superiority for OOD generalization on medical images \cite{Zhou:22}. Therefore, we introduce the \emph{WIN} which injects noises into each instance instead of slices of batch.

Specifically, the perturbation (i.e., noise injection) for each instance is conducted with the feature standardization operation. We standardize the feature $f$ with approximations of global statistics (i.e., $\mu_{IN}$ and $\sigma_{IN}^{2}$) (see Eq. \ref{eq:Standardization}). As shown in Fig. \ref{fig:WIN} (c), the mean and variance for each channel and instance are calculated within a random window. Formally, given a window that is specified by the top-left coordinate $(\bar{x},\bar{y})$ and bottom-right coordinate $(\tilde{x},\tilde{y})$, the mean and variance, $\mu_{WIN}$, $\sigma_{WIN}^{2}$ $\in \mathbb{R}^{N\times C}$ are computed as:
\begin{equation}
\begin{aligned}
&\mu_{WIN}^{n,c}=\frac{1}{\bar{H} \bar{W}} \sum_{h=\bar{x}}^{\tilde{x}} \sum_{w=\bar{y}}^{\tilde{y}} f_{n, c, h, w}, \\
&{\sigma_{WIN}^{2}}^{n,c}=\frac{1}{\bar{H} \bar{W}} \sum_{h=\bar{x}}^{\tilde{x}} \sum_{w=\bar{y}}^{\tilde{y}}{\left(f_{n, c, h, w}-\mu_{WIN}^{n,c}\right)}^{2}
\end{aligned}
\end{equation}
where $\bar{H}$ and $\bar{W}$ are the height and width of a window. Algorithm \ref{algorithm} demonstrates the \emph{Window} sampling strategy: we repeat generating a window denoted as $\{x, y, \bar{W}, \bar{H}\}$ and placing this window on the feature until the size of the random window exceeds $\tau \times H \times W$. $\tau$ is a threshold for the ratio of window size. LCN \cite{ortiz2020local} also uses the statistics of windows, but they normalize each feature based on the statistics of its local neighborhoods and corresponding feature group. Besides, their prime use is in dense prediction tasks such as object detection, semantic segmentation, and instance segmentation.

Considering such perturbation is marginal for the images with consistent background (e.g., chromosome images), we propose another strategy called \emph{Block} which computes the $\mu_{WIN}$ and $\sigma_{WIN}^{2}$ within multiple small windows. Firstly, we divide the input feature $f$ into $B$ non-overlapping blocks. Each block corresponds to a particular patch on the input image regardless of the scale of feature. At a certain scale $s$, the $i$-th block is denoted as $\{\bar{x}^{s}_{i},\bar{y}^{s}_{i},\tilde{x}^{s}_{i},\tilde{y}^{s}_{i}\}$. For example, the resolution of input images is $224 \times 224$ and the patch size is $32 \times 32$, so they will be divided into $49$ blocks regardless of the feature scales. Then, we sample $\tau B$ random blocks without replacement, following a uniform distribution. The mean and variance $\mu_{WIN}$, $\sigma_{WIN}^{2}$ at the scale $s$ are computed as:

\begin{equation}
\begin{aligned}
&\mu_{WIN}^{n,c}=\frac{1}{(\tau B )(s^{2} \hat{H} \hat{W} )} \sum_{i}^{} \sum_{h=\bar{x}^{s}_{i}}^{\tilde{x}^{s}_{i}} \sum_{w=\bar{y}^{s}_{i}}^{\tilde{y}^{s}_{i}} f_{n, c, h, w}, \\
&{\sigma_{WIN}^{2}}^{n,c}=\frac{1}{(\tau B)(s^{2} \hat{H} \hat{W} )} \sum_{i}^{}\sum_{h=\bar{x}^{s}_{i}}^{\tilde{x}^{s}_{i}} \sum_{w=\bar{y}^{s}_{i}}^{\tilde{y}^{s}_{i}}{\left(f_{n, c, h, w}-\mu_{WIN}^{n,c}\right)}^{2}
\end{aligned}
\end{equation}
where $i$ is the index of random blocks and $\hat{H}$ and $\hat{W}$ are the original height and width of the patches. This strategy avoids the zero variance (e.g., a window on the consistent background) effectively.

In order to diversify the perturbations and smooth the model response, \emph{WIN} mixes the local statistics and the global statistics. The mixing statistics can be formulated as follows:
\begin{equation}
\begin{aligned}
&\mu=\lambda\mu_{WIN}+(1-\lambda)\mu_{IN}, \\
&\sigma^{2}=\lambda\sigma_{WIN}^{2}+(1-\lambda)\sigma_{IN}^{2},
\end{aligned}
\end{equation}
where $\lambda \in \mathbb{R}^{N\times C}$ is a random instance-specific weight sampled from a beta distribution $ Beta(\alpha, \alpha)$.

To keep the evaluation deterministic, we adopt $\mu=\mu_{IN}$ and $\sigma^{2}=\sigma_{IN}^{2}$ during evaluation as the mixing statistics are approximations of global statistics (see Fig. \ref{fig:WIN}). The mixing statistics are equivalent to global statistics when $\tau=1.0$ or $\lambda=1.0$.

\subsection{Training and evaluation discrepancy}
Normalizing the features with \emph{WIN} can regularize the CNNs training effectively and improve the OOD generalization significantly. However, the evaluation statistics cannot be strictly aligned to training statistics and the model parameters are trained to fit features normalized by $\mu_{WIN}$, $\sigma_{WIN}^{2}$, which degrades the model generalization. On the other hand, normalizing the feature with mixing statistics (i.e., training mode) or global statistics (i.e., evaluation mode) can produce two correlated views of the same sample. Minimizing the discrepancy between training and evaluation is a natural self-learning task to incorporate the invariance among different views \cite{gao2021simcse, liang2021r-drop}.

To these ends, we introduce the \emph{WIN-WIN} for classification tasks, which needs twice forward passes. As shown in Fig. \ref{fig:WIN}(d), given the input $x$ with a ground-truth label $y$, the first pass uses mixing statistics to normalize the features and outputs a logits $\hat{y}$, while the second pass uses global statistics to normalize the features and outputs a logits $\bar{y}$. \emph{WIN-WIN} will compel the model to minimize the Jensen-Shannon divergence between $\hat{y}$ and $\bar{y}$ and the cross-entropy between $\hat{y}$ and $y$ and $\bar{y}$ and $y$. The Jensen-Shannon divergence loss can be written as:
\begin{equation}
\begin{aligned}
\mathcal{L}_{JSD}&=\frac{1}{2}(\mathcal{D}_{K L}(softmax(\hat{y}) \| softmax(\bar{y}))\\
&+ \mathcal{D}_{K L}(softmax(\bar{y}) \| softmax(\hat{y}))) 
\\
\end{aligned}
\end{equation}
where the logits are converted to a probability vector via $softmax$ function. Meanwhile, the cross-entropy loss can be written as:
\begin{equation}
\begin{aligned}
\mathcal{L}_{CE}&=\frac{1}{2}(\mathcal{H}(softmax(\hat{y}),y)+\mathcal{H}(softmax(\bar{y}), y))). 
\end{aligned}
\end{equation}
Finally, the total loss for classification tasks can be written as:
\begin{equation}
\begin{aligned}
\label{eq:total}
\mathcal{L}_{Total}&=\mathcal{L}_{CE}+\delta\mathcal{L}_{JSD}
\end{aligned}
\end{equation}
where $\delta$ is used to balancing $\mathcal{L}_{CE}$ and $\mathcal{L}_{JSD}$.

\section{Experiments}
\subsection{Datasets and Evaluation Metrics}
\begin{figure}[tb]
    \begin{center}
    \includegraphics[width=\linewidth]{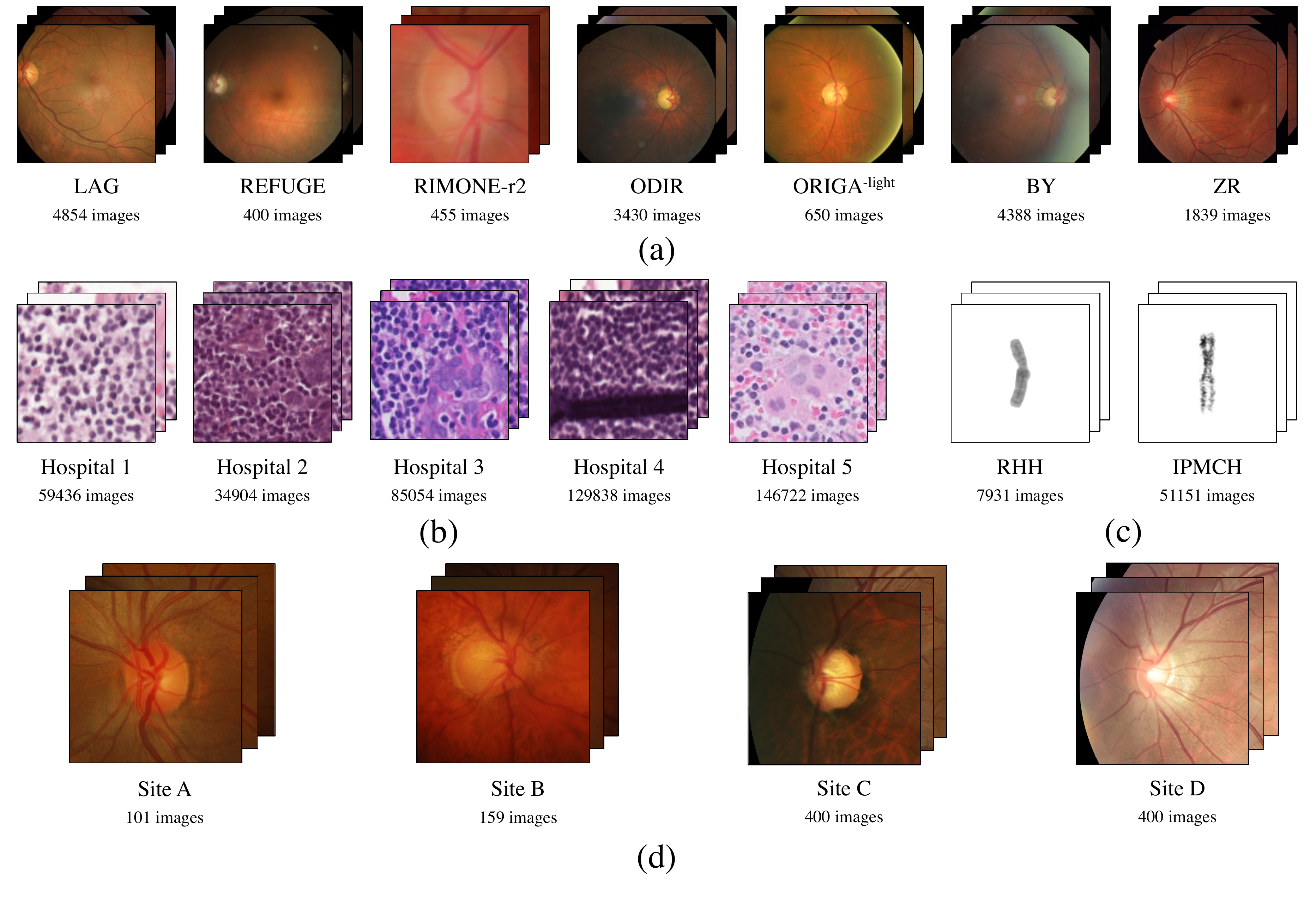}
    \vspace{-0.9cm}
    
    \caption{Example cases and sample numbers of each data source in our experiments. (a) glaucoma detection, (b) breast cancer detection, (c) chromosome classification, and (d) optic disc and cup segmentation} 
    \label{fig:data} 
    \end{center}
    \vspace{-0.7cm}
    
\end{figure}

To demonstrate the generality of our methods, the experiments covered a wide range of tasks: binary classification, multiclass classification, and image segmentation. And, the datasets range from common OOD generalization benchmarks to real-world applications. A noticeable domain gap is existing in the image appearance and image quality (see Fig. \ref{fig:data}). 

\textbf{Binary Classification.} Following \cite{Zhou:22}, we set up a benchmark using seven glaucoma detection datasets from seven sites collected with various scanners, including five public datasets (i.e., LAG, REFUGE, RIMONE-r2, ODIR, and ORIGA$\rm ^{-light}$ ) and two private datasets (i.e., BY and ZR). BY and ZR are collected from Peking University Third Hospital and the Second Affiliated Hospital of Zhejiang University, respectively. Furthermore, the experiments were also conducted on an OOD generalization benchmark Camelyon17. Camelyon17 is a breast cancer detection benchmark of WILDS \cite{koh2021wilds}, which includes five data sources. All images in Camelyon17 were resized from $96\times96$ to $256\times256$ to facilitate the window sampling. Fig. \ref{fig:data} (a) and (b) show two benchmarks' example cases and sample numbers. Here, the evaluation metric we applied was the mean clean area under the curve ($m$-$cAUC$) following \cite{Zhou:22}. Each dataset was divided into $80\%$ and $20\%$ for training and validation, while other datasets were held out for testing. Before aggregating, each test dataset's result (i.e., the area under the curve) will be recalibrated by its inherent difficulties. A slight difference to \cite{Zhou:22} is that we only run it once, taking into account multiple test datasets and time-saving. Higher $m$-$cAUC$ means better OOD generalization.

\textbf{Multiclass Classification.} As shown in Fig. \ref{fig:data} (c), we collected chromosome images from two hospitals using different microscopes to cross-evaluate the OOD generalization performance in chromosome classification, including $7931$ samples from The Obstetrics Gynecology Hospital of Fudan University (RHH) and $51151$ samples from The International Peace Maternity Child Health Hospital of China welfare institute (IPMCH). The dataset from one hospital was divided into a training set and a validation set according to the $80\%:20\%$ proportion and another hospital data was used for testing. The result was reported in Top-1 accuracy ($Acc.$). Besides, we also conducted experiments on two robustness benchmarks CIFAR-10-C and CIFAR-100-C \cite{DBLP:conf/iclr/HendrycksD19} and a commonly used domain generalization (DG) benchmark Digit-DG \cite{KaiyangZhou2020LearningTG}. CIFAR-10-C and CIFAR-100-C consist of $75$ corrupted versions of the original test set. The metric for them is the mean corruption error ($mCE$) across all $15$ corruptions and $5$ severities for each corruption. Lower is better. Digit-DG consists of four handwritten digit recognition datasets (MNIST, MNIST-M, SVHN, and SYN) with distribution shifts for font style, stroke color, and background. \emph{Following prior DG works \cite{zhou2021mixstyle}, we applied the leave-one-domain-out protocol for Digit-DG and reported the Top-1 accuracy on each target dataset}. Chromosome images, CIFAR-10-C, and CIFAR-100-C were resized to $256\times256$ and the images of Digit-DG were resized to $224\times224$.

\textbf{Image Segmentation.} We employed the retinal fundus images in \cite{ShujunWang2020DoFEDF} for optic cup and disc (OC/OD) segmentation. As shown in Fig. \ref{fig:data} (c), they are collected from $4$ sites using different scanners\cite{ShujunWang2020DoFEDF}, and each image was center-cropped and resized to $384 \times 384$ as network inputs. We randomly divided the images on each site into training and test sets following the proportion of $80\%:20\%$. In our experiments, the training sets of three sites (e.g., Site A, B, and C) were used for training, and the test set of remaining site (e.g., Site D) was used for testing. \emph{It is a common setting of DG}. The segmentation result was reported in the Dice coefficient ($Dice$). Higher is better.


\begin{table*}[tb]
    \caption{Comparison on the glaucoma detection task of \emph{WIN} and other state-of-the-art methods. Experimental results are reported with the ResNet-50  for $240$ epochs  on LAG. \emph{DeepAll} serves as the upper bound for each dataset. The performance on each dataset is evaluated with the area under the curve ($AUC$). Top results are bolded.}
    \label{tab:glaucoma}
    \vspace{-0.2cm}
    \centering
    \begin{tabular}{lcccccccc}
    \toprule
    \multirow{2}{*}{Methods} &  IND ($AUC$) & \multicolumn{6}{c}{OOD ($AUC$)} & $m$-$cAUC$ $\uparrow$ \\
    \cmidrule(lr){3-8}
      &  LAG & REFUGE$\quad$ & RIMONE-r2 &  ODIR$\quad$ & ORIGA$\rm$ $^{\small{-light}}$  & BY$\qquad$ & ZR$\qquad$  & (mean$\pm$ std)  \\
    \midrule
    \emph{DeepAll}  & 0.982 &  0.989 & 0.912 & 0.874 & 0.782 & 0.989 & 0.999 & 1.000$\pm$0.00 \\
    \midrule
    \emph{BN} \cite{DBLP:conf/icml/IoffeS15} & 0.991 &  0.740 & 0.390 & 0.596 & 0.698 & 0.597 & 0.771 & 0.688$\pm$0.15 \\
    \emph{IN} \cite{DBLP:journals/corr/UlyanovVL16} & 0.983 &  0.870 & 0.651 & 0.773 & 0.741 & 0.763  & 0.841 & 0.840$\pm$0.08 \\
    \emph{WIN} & 0.984 &  0.874 & 0.668 & 0.800 & 0.770 & 0.836  & 0.878 & 0.873$\pm$0.08 \\
    \emph{WIN-WIN} & 0.989 &  \textbf{0.896} & \textbf{0.691} & \textbf{0.820} & \textbf{0.775} & \textbf{0.874}  & 0.879 & \textbf{0.893$\pm$0.07} \\
    \midrule
    LCN \cite{ortiz2020local} & 0.988 &  0.820 & 0.578 & 0.698 & 0.675 & 0.712  & 0.771 & 0.769$\pm$0.08 \\
    SNR \cite{jin2021style} & 0.993 &  0.806 & 0.442 & 0.609 & 0.660 & 0.563  & 0.779 & 0.698$\pm$0.13 \\
    BIN \cite{nam2018batch} & 0.984 &  0.875 & 0.560 & 0.771 & 0.760 & 0.819  & \textbf{0.889} & 0.845$\pm$0.11  \\
    SN \cite{luo2019switchable} & 0.992 & 0.791 & 0.621 & 0.629 & 0.644 & 0.591  & 0.751 & 0.729$\pm$0.08 \\
    IBN-a \cite{DBLP:conf/eccv/PanLST18} & 0.993 & 0.777 & 0.597 & 0.667 & 0.728 & 0.550  & 0.795 & 0.748$\pm$0.12 \\
    IBN-b \cite{DBLP:conf/eccv/PanLST18} & 0.991 & 0.862 & 0.700 & 0.730 & 0.706 & 0.637  & 0.811 & 0.805$\pm$0.08 \\
    GN \cite{wu2018group} & 0.991 & 0.837 & 0.499 & 0.724 & 0.683 & 0.661  & 0.821 & 0.764$\pm$0.12 \\
    RBN \cite{gao2021representative} & 0.991 & 0.755 & 0.458 & 0.617 & 0.724 & 0.633  & 0.838 & 0.729$\pm$0.14 \\
    CNSN \cite{tang2021crossnorm} & 0.987 & 0.750 & 0.459 & 0.582 & 0.617 & 0.503  & 0.592 & 0.636$\pm$0.11 \\
    \midrule
    AutoAugment \cite{cubuk2019autoaugment} & 0.996 & 0.605 & 0.452 & 0.548 & 0.646 & 0.601  & 0.761 & 0.655$\pm$0.11 \\
    RandAugment \cite{cubuk2020randaugment} & 0.992 & 0.712 & 0.462 & 0.650 & 0.761 & 0.771  & 0.779 & 0.750$\pm$0.14 \\
    Patch Gaussian \cite{lopes2019improving} & 0.995 & 0.653 & 0.428 & 0.562 & 0.614 & 0.531  & 0.606 & 0.617$\pm$0.10 \\
    randomErasing \cite{zhong2020random} & 0.994 & 0.700 & 0.401 & 0.526 & 0.627 & 0.540  & 0.660 & 0.626$\pm$0.12 \\
    \midrule
    mixup \cite{DBLP:conf/iclr/ZhangCDL18}  & 0.993 & 0.605 & 0.443 & 0.583 & 0.640 & 0.520  & 0.746 & 0.643$\pm$0.12 \\
    CutMix \cite{yun2019cutmix} & 0.995 & 0.679 & 0.455 & 0.590 & 0.683 & 0.675  & 0.768 & 0.697$\pm$0.11 \\
    AugMix \cite{DBLP:conf/iclr/HendrycksMCZGL20} & 0.995 & 0.679 & 0.455 & 0.590 & 0.683 & 0.675  & 0.768 & 0.697$\pm$0.11 \\
    PixMix \cite{hendrycks2022robustness} & \textbf{0.997} & 0.702 & 0.423 & 0.640 & 0.693 & 0.708  & 0.783 & 0.715$\pm$0.13 \\
    \midrule
    mainfold mixup \cite{verma2019manifold} & 0.993 & 0.637 & 0.389 & 0.502 & 0.653 & 0.664  & 0.348 & 0.583$\pm$0.16 \\
    MixStyle \cite{zhou2021mixstyle} & 0.995 & 0.705 & 0.364 & 0.602 & 0.678 & 0.590  & 0.829 & 0.682$\pm$0.16 \\
    \bottomrule
    \end{tabular}
    \vspace{-0.4cm}
\end{table*}

\begin{table*}[tb]
    \caption{Comparison with baselines on the breast cancer detection task. Experimental results are reported with the ResNet-50. \emph{DeepAll} serves as the upper bound for each dataset. The performance on each dataset is evaluated with the area under the curve ($AUC$). Top results are bolded. The in-distribution result is underlined.}
    \label{tab:breast}
    \vspace{-0.2cm}
    \centering
    \begin{tabular}{lccccccccc}
    \toprule
    \multirow{2}{*}{Methods} & Training & \multirow{2}{*}{epochs}  & \multicolumn{5}{c}{$AUC$} & $m$-$cAUC$ $\uparrow$ & \multirow{2}{*}{Avg. ($m$-$cAUC$ $\uparrow$)} \\
    \cmidrule(lr){4-8}
    & data & & H1 & H2 & H3 & H4 & H5 & (mean$\pm$std)  \\
    
    \midrule
    \emph{DeepAll} & All & 10 & 0.998 & 0.997 & 0.999 & 0.999 & 0.999 & 1.000$\pm$0.00 &  1.000 \\
    \midrule
    \multirow{5}{*}{\emph{BN} \cite{DBLP:conf/icml/IoffeS15}} & H1 & 50 & \underline{0.999} & 0.868 & 0.696 & 0.908 & 0.528 & 0.752$\pm$0.18 &\multirow{5}{*}{0.659} \\
     & H2 & 50 & 0.721 & \underline{0.999} & 0.184 & 0.702 & 0.565 & 0.544$\pm$0.25 \\
     & H3 & 40 & 0.718 & 0.844 & \underline{1.000} & 0.327 & 0.553 & 0.612$\pm$0.22 \\
     & H4 & 25 & 0.730 & 0.697 & 0.410 & \underline{1.000} & 0.602 & 0.611$\pm$0.14 \\
     & H5 & 25 & 0.767 & 0.708 & 0.668 & 0.946 & \underline{1.000} & \textbf{0.774$\pm$0.12} \\
    \midrule
    \multirow{5}{*}{\emph{IN} \cite{DBLP:journals/corr/UlyanovVL16}} & H1 & 50 & \underline{0.999} & 0.936 & 0.816 & 0.967 & 0.867 & 0.898$\pm$0.07 &\multirow{5}{*}{0.758}  \\
     & H2 & 50 & 0.952 & \underline{0.998} & 0.949 & 0.908 & 0.499 & 0.828$\pm$0.22\\
     & H3 & 40 & 0.726 & 0.544 & \underline{0.999} & 0.434 & 0.597 & 0.576$\pm$0.12 \\
     & H4 & 25 & 0.957 & 0.886 & 0.664 & \underline{0.999} & 0.773 & \textbf{0.822$\pm$0.13} \\
     & H5 & 25 & 0.621 & 0.680 & 0.771 & 0.580 & \underline{0.999} & 0.664$\pm$0.08\\
    \midrule
    \multirow{5}{*}{\emph{WIN}} & H1 & 50 & \underline{1.001} & 0.926 & 0.878 & 0.974 & 0.891 & 0.917$\pm$0.04 &\multirow{5}{*}{0.757} \\
     & H2 & 50 & 0.970 & \underline{1.000} & 0.933 & 0.959 & 0.647 &  \textbf{0.877$\pm$0.15}\\
     & H3 & 40 & 0.634 & 0.496 & \underline{1.000} & 0.465 & 0.678 & 0.569$\pm$0.10 \\
     & H4 & 25 & 0.943 & 0.897 & 0.503 & \underline{1.000} & 0.761 & 0.776$\pm$0.20 \\
     & H5 & 25 & 0.629 & 0.665 & 0.692 & 0.599 & \underline{0.999} & 0.647$\pm$0.41\\
    \midrule 
    \multirow{5}{*}{\emph{WIN-WIN}} & H1 & 50 & \underline{0.998} & 0.944 & 0.935 & 0.973 & 0.942 & \textbf{0.950$\pm$0.02} &\multirow{5}{*}{\textbf{0.778}} \\
     & H2 & 50 & 0.967 & \underline{0.997} & 0.918 & 0.966 & 0.573 &  0.857$\pm$0.19\\
     & H3 & 40 & 0.734 & 0.504 & \underline{0.999} & 0.461 & 0.795 & \textbf{0.625$\pm$0.17} \\
     & H4 & 25 & 0.921 & 0.892 & 0.506 & \underline{0.999} & 0.629 & 0.738$\pm$0.20 \\
     & H5 & 25 & 0.665 & 0.800 & 0.796 & 0.609 & \underline{0.999} & 0.719$\pm$0.10\\
    \bottomrule
    \end{tabular}
    \vspace{-0.7cm}
\end{table*}

\subsection{Implementation Details}
We conducted all experiments on two NVIDIA GeForce GTX 2080Ti with PyTorch implementation. All models were trained from scratch using a cross-entropy loss. The training epoch is determined by tasks. The hyper-parameters of Beta distribution used in mixing statistics were empirically set to $\alpha=0.1$. And, we set the ratio threshold  $\tau=0.7$ and $\delta=0.3$. With the exception of chromosome classification, all tasks used the \emph{Window} strategy. The \emph{Block} strategy was adopted for chromosome classification. More detailed settings are listed as follows.

\textbf{Binary Classification.} We optimized the ResNet-50 \cite{DBLP:conf/cvpr/HeZRS16} with the following settings: SGD optimizer with Nesterov momentum of $0.9$; weight decay of $1e-5$; batch size of $64$. The learning rate increased linearly from $0$ to $3e-3$ at the first $5$ epochs and linearly decayed to $0$ following a cosine decay schedule. The image was horizontally flipped  with a $50\%$ probability and randomly cropped to $224\times224$ before feeding into the model. 

\textbf{Multiclass classification.} For the chromosome classification task, we ran five times with the same setting of \textbf{Binary Classification}. The CIFAR-10-C and CIFAR-100-C experiments were built based on \textsf{PIXMIX} \cite{hendrycks2022robustness}\footnote{https://github.com/andyzoujm/pixmix}. SGD with a Nesterov momentum of $0.9$ was adopted as the optimizer to train the ResNet-18 \cite{DBLP:conf/cvpr/HeZRS16}. The batch size was set to $64$. The learning rate was set to $0.3$ and decayed following a cosine learning rate schedule. The augmentation policy was the same as \textbf{Binary Classification}. In the experiments of Digit-DG, five ResNet-18 models were optimized for $180$ epochs with the following settings: SGD optimizer with a momentum of $0.9$; weight decay of $5e-4$; batch size of $64$; cosine decay schedule. The code was built based on \textsf{mixstyle-release} \cite{zhou2021mixstyle} \footnote{https://github.com/KaiyangZhou/mixstyle-release}. There was no augmentation for the input images.

\textbf{Image segementaion.} We built the code based on \textsf{DoFE} \cite{ShujunWang2020DoFEDF}\footnote{https://github.com/emma-sjwang/Dofe} and \textsf{MONAI} \footnote{https://github.com/Project-MONAI/MONAI}. We trained the $5$ layer U-Net \cite{OlafRonneberger2015UNetCN} for five times and optimized the cross-entropy loss with the Adam optimizer. The learning rate was set to $1e-3$ in the first $60$ epochs and linearly decreased to $2e-4$ in the last $20$ epochs. $\beta_{1}$ and $\beta_{2}$ were set to $0.9$ and $0.99$, respectively. The input images were randomly cropped and resized to $256\times256$.


\begin{table}[tb]
    \caption{Comparison with baselines on the chromosome classification. Experimental results are reported with the ResNet-50 and averaged over five runs. The results are reported in Top-1 accuracy ($Acc.$,\%). Top results are bolded. $e.$ is an abbreviation of epochs.}
    \label{tab:chromosome}
    \vspace{-0.2cm}
    \centering
    \begin{tabular}{lcccc}
    \toprule
    \multirow{2}{*}{ Methods} & \multicolumn{2}{c}{RHH $\rightarrow$ IPMCH (200 $e.$)}  & \multicolumn{2}{c}{IPMCH $\rightarrow$ RHH  (60 $e.$) }    \\
    \cmidrule(lr){2-3} \cmidrule(lr){4-5}
    & IND  & OOD  & IND  & OOD  \\
    \midrule
    \emph{BN} \cite{DBLP:conf/icml/IoffeS15} &\textbf{93.3$\pm$0.3} & 21.4$\pm$1.4 & \textbf{97.6$\pm$0.1} & 24.8$\pm$1.2 \\
    \emph{IN} \cite{DBLP:journals/corr/UlyanovVL16} & 92.6$\pm$0.2 & 35.2$\pm$2.7 &97.5$\pm$0.1 & 31.9$\pm$1.9  \\
    \emph{WIN} & 92.4$\pm$0.3 & 41.0$\pm$3.5 & 97.7$\pm$0.2 & 32.2$\pm$1.9  \\
    
    \emph{WIN-WIN} & 92.9$\pm$0.3 & \textbf{43.4$\pm$2.2} & 97.6$\pm$0.2 & \textbf{32.5$\pm$1.7}  \\

    \bottomrule
    \end{tabular}
    \vspace{-0.8cm}
\end{table}

\begin{table}[htb]
    \caption{Comparison with baselines on the CIFAR-10-C and CIFAR-100-C. Experimental results are reported with the ResNet-50 trained by the original CIFAR-10 or CIFAR-100. IND performance and OOD performance are reported in Top-1 Accuracy ($Acc.\uparrow$,\%) and mean corruption error ($mCE \downarrow$, \%), respectively. Top results are bolded.}
    \label{tab:cifar}
    \vspace{-0.2cm}
    \centering
    \begin{tabular}{lcccc}
    \toprule
    \multirow{2}{*}{ Methods} & \multicolumn{2}{c}{CIFAR-10-C (180 $e.$)}  & \multicolumn{2}{c}{CIFAR-100-C (200 $e.$)} \\
    \cmidrule(lr){2-3} \cmidrule(lr){4-5}
    & IND  & OOD  & IND   & OOD \\
    \midrule
    \emph{BN} \cite{DBLP:conf/icml/IoffeS15} & \textbf{95.6} & 23.5 & 75.3 & 50.6 \\
    \emph{IN} \cite{DBLP:journals/corr/UlyanovVL16} & 94.6 & 18.5 & 75.1 & 48.4 \\
    \emph{WIN} & 94.2 & 18.1 & 75.2 & 46.4 \\
    \emph{WIN-WIN} & 94.6 & \textbf{17.5} & \textbf{76.4} & \textbf{45.7} \\

    \bottomrule
    \end{tabular}
    \vspace{-0.5cm}
\end{table}

\begin{table}[htb]
    \caption{Comparison with baselines on the Digit-DG. Experimental results are reported with the ResNet-18 and averaged over five random splits. The performance on the target domain is evaluated with Top-1 accuracy ($Acc.$,\%). M: MNIST. MNIST-M: MM. SVHN: SV. SYN: SY. Top results are bolded.}
    \label{tab:digit-dg}
    \vspace{-0.2cm}
    \centering
    \begin{tabular}{lccccc}
    \toprule
    \multirow{2}{*}{ Methods} & \multicolumn{4}{c}{Target domain} & \multirow{2}{*}{ Avg.} \\
    \cmidrule(lr){2-5}

    &M&MM& SV&SY& \\
     \midrule
    \emph{BN} \cite{DBLP:conf/icml/IoffeS15} & 93.8$\pm$0.5 & 45.0$\pm$0.9 & 55.2$\pm$1.7 & 83.7$\pm$0.8 & 69.4 \\ 
    \emph{IN} \cite{DBLP:journals/corr/UlyanovVL16} & 95.1$\pm$0.3 & 48.8$\pm$2.5 & 52.8$\pm$0.7 & 86.5$\pm$0.6 & 70.8\\
    \emph{WIN} & 97.0$\pm$0.3 & 53.1$\pm$1.0 & 59.1$\pm$1.8 & 89.2$\pm$0.7  & 74.4\\ 
    \emph{WIN-WIN} & \textbf{97.6$\pm$0.1} & \textbf{62.5$\pm$0.8} & \textbf{66.0$\pm$1.6} & \textbf{92.7$\pm$0.3}  & \textbf{79.7}\\ 
    \bottomrule
    \end{tabular}
    \vspace{-0.7cm}
\end{table}

\subsection{Comparison with Baseline}
\label{sec:baseline}

Two baseline methods \emph{BN} and \emph{IN} are introduced since they are the most popular normalization methods. Both of them have demonstrated their effectiveness in improving model generalization. In practice, as we conducted the experiments on two GPUs, the \emph{BN} is equivalent to GBN \cite{dimitriou2020new}. In the following, we first compare \emph{WIN} with \emph{BN} and \emph{IN}. Then, we compare \emph{WIN-WIN} with these normalization techniques.

For the binary classification tasks, we aggregated all datasets of the same task and trained the \emph{DeepAll}. \emph{DeepAll} provides a difficulty coefficient and can be served as the upper bound for each dataset \cite{Zhou:22}. Table \ref{tab:glaucoma} and \ref{tab:breast} present the experimental results of two tasks. Compared with \emph{BN}, \emph{WIN} surpassed it for the OOD generalization remarkably (the $m$-$cAUC$ of $+0.185$ on glaucoma detection and the averaged $m$-$cAUC$ of $+0.098$ on breast cancer detection). Compared with \emph{IN}, \emph{WIN} led it with $+0.033$ $m$-$cAUC$ on glaucoma detection and approached it with $-0.001$ averaged $m$-$cAUC$ on breast tumor detection. The superiority of \emph{WIN} is more obvious for small training datasets (e.g., H1 and H2). Besides, their results on IND data revealed a slight variation. It is insensitive to normalization methods. Taking advantage of \emph{WIN}, \emph{WIN-WIN} achieved the best results on two tasks, which improved \emph{WIN} by $+0.020$ $m$-$cAUC$ on glaucoma detection and $+0.021$ averaged $m$-$cAUC$ on breast tumor detection.

Multiclass classification is a more challenging task. The superiority of our methods still exists. Table \ref{tab:chromosome} presents the experimental results of chromosome classification. This task asks the models to classify the input chromosome image into 24 types. Since a few images in RHH are mislabeled by beginners, the distribution shifts appear in the input images and the labels simultaneously. In Table \ref{tab:chromosome}, RHH$\rightarrow$IPMCH denotes training on RHH and testing on IPMCH, and IPMCH$\rightarrow$RHH reverses this setting. The accuracy of \emph{WIN} excelled that of \emph{BN} by $19.6\%$ and that of \emph{IN} by $5.8\%$ in RHH$\rightarrow$IPMCH. On another setting, \emph{WIN} was $+7.4\%$ for \emph{BN} and $+0.3\%$ for \emph{IN}. Meanwhile, \emph{WIN-WIN} improved \emph{WIN} by $+2.4\%$ in RHH$\rightarrow$IPMCH and $+0.3\%$ in IPMCH$\rightarrow$RHH. Furthermore, we verified our methods on the common benchmark CIFAR-10-C, CIFAR-100-C, and Digit-DG. As shown in Table \ref{tab:cifar}, our methods effectively improved the corruption robustness compared with the baselines. In the experiments of Digit-DG, three domains were used for training and one domain for testing. Our methods greatly outperformed the baselines regardless of the target domain (see Table \ref{tab:digit-dg}). \emph{WIN} and \emph{WIN-WIN} excelled the \emph{IN} by the averaged Top-1 accuracy of $+3.6\%$ and $+8.9\%$, respectively. It strongly demonstrates that our methods can improve the OOD generalization with multi-source domain data.

\begin{table*}[htb]
    \caption{Comparison with baselines on OC/OD segmentation task. Experimental results are reported with a U-Net and averaged over five runs. The performance on each test dataset was evaluated with a dice score ($Dice\uparrow$). Top results are bolded.}
    \label{tab:segmentation}
    \vspace{-0.2cm}
    \centering
    \begin{tabular}{lccccccccc}
    \toprule
    \multirow{2}{*}{ Methods} & \multicolumn{2}{c}{Site A}  & \multicolumn{2}{c}{Site B} & \multicolumn{2}{c}{Site C} & \multicolumn{2}{c}{Site D} & \multirow{2}{*}{ Avg. ($Dice \uparrow$)} \\
    \cmidrule(lr){2-3} \cmidrule(lr){4-5} \cmidrule(lr){6-7} \cmidrule(lr){8-9} 
    & OC  & OD  & OC  & OD & OC  & OD & OC  & OD & \\
    \midrule
    \emph{BN} \cite{DBLP:conf/icml/IoffeS15} & 0.709$\pm$0.05 & 0.921$\pm$0.01 & 0.666$\pm$0.02& 0.834$\pm$0.01 & 0.779$\pm$0.01 & 0.906$\pm$0.01 & 0.813$\pm$0.01 & 0.905$\pm$0.02 & 0.817 \\
    \emph{IN} \cite{DBLP:journals/corr/UlyanovVL16} & 0.753$\pm$0.03 & 0.948$\pm$0.00 & 0.710$\pm$0.02 & \textbf{0.867$\pm$0.01} & 0.819$\pm$0.02 & 0.938$\pm$0.00 & \textbf{0.848$\pm$0.02} & \textbf{0.927$\pm$0.00} & 0.851\\
    \emph{WIN} & \textbf{0.756$\pm$0.03} & \textbf{0.952$\pm$0.00} & \textbf{0.734$\pm$0.01} & 0.848$\pm$0.01 & \textbf{0.839$\pm$0.01} & \textbf{0.940$\pm$0.01} & 0.846$\pm$0.01 & 0.924$\pm$0.00 & \textbf{0.855} \\

    \bottomrule
    \end{tabular}
    \vspace{-0.3cm}
\end{table*}

In addition to the classification tasks, we investigated the application of \emph{WIN} on image segmentation. Table \ref{tab:segmentation} presents that \emph{WIN} outperformed \emph{BN} with an average of $+0.038$ dice score and outperformed \emph{IN} with an average of $+0.004$ dice score on the OC/OD segmentation. This demonstrates once again that our method can improve the OOD generalization with multi-source domain data.
 
In summary, \emph{WIN} is a versatile normalization technique for improving the OOD generalization. It significantly surpasses the \emph{BN} and \emph{IN} on a number of tasks regardless of training setting (i.e., single source domain or multi-source domains). On this basis, \emph{WIN-WIN} can further improve it for OOD generalization on classification tasks.

\subsection{Ablation Analysis}
\begin{table}[tb] 
    \caption{Ablation study on glaucoma detection using ResNet-50. }
    \vspace{-0.2cm}
    \label{tab:ablation}
    \centering
    \begin{tabular}{lccccc}
    \toprule
    Methods & Strategies & Stat. & Mixing & Consist. & $m$-$cAUC$$\uparrow$ \\
    \midrule
    \multirow{8}{*}{ \emph{WIN}}   & \emph{Window}     & $\mu$,$\sigma$ & \checkmark & \texttimes & 0.873  \\
       & \emph{Window}     & $\mu$,$\sigma$ & \texttimes & \texttimes & 0.871 \\
       & \emph{Window}     & $\mu$        & \texttimes & \texttimes & 0.704 \\
       & \emph{Window}     & $\sigma$     & \texttimes & \texttimes & 0.711 \\
       & \emph{Global (IN)}     & $\mu$,$\sigma$ & \texttimes & \texttimes & 0.840 \\
       & \emph{Block}      & $\mu$,$\sigma$ & \texttimes & \texttimes & 0.846 \\
       & \emph{Pixel}     & $\mu$,$\sigma$ & \texttimes & \texttimes & 0.814  \\
       & \emph{Mask}      & $\mu$,$\sigma$ & \texttimes & \texttimes & 0.779  \\
       & \emph{Speckle}   & $\mu$,$\sigma$ & \texttimes & \texttimes & 0.813  \\
    
    \midrule
    \multirow{3}{*}{ \emph{WIN-WIN}} & \emph{Window}     & $\mu$,$\sigma$ & \checkmark & logits & \textbf{0.893} \\
     
     & \emph{Window}     & $\mu$,$\sigma$ & \texttimes & logits     & 0.876 \\
     & \emph{Window}     & $\mu$,$\sigma$ & \checkmark & features   & 0.867 \\
    
    \bottomrule
    \end{tabular}
    \vspace{-0.7cm}
\end{table}

In this section, we looked at the options for $\mu$ and $\sigma$ in \emph{WIN} first. After that, we investigated the effect of the mechanisms behind \emph{WIN-WIN}. Finally, we presented the analysis of hyper-parameter sensitivity. All experiments in this section were conducted using a ResNet-50 trained with LAG.

\begin{figure}[tb]
    \begin{center}
    \includegraphics[width=0.93\linewidth]{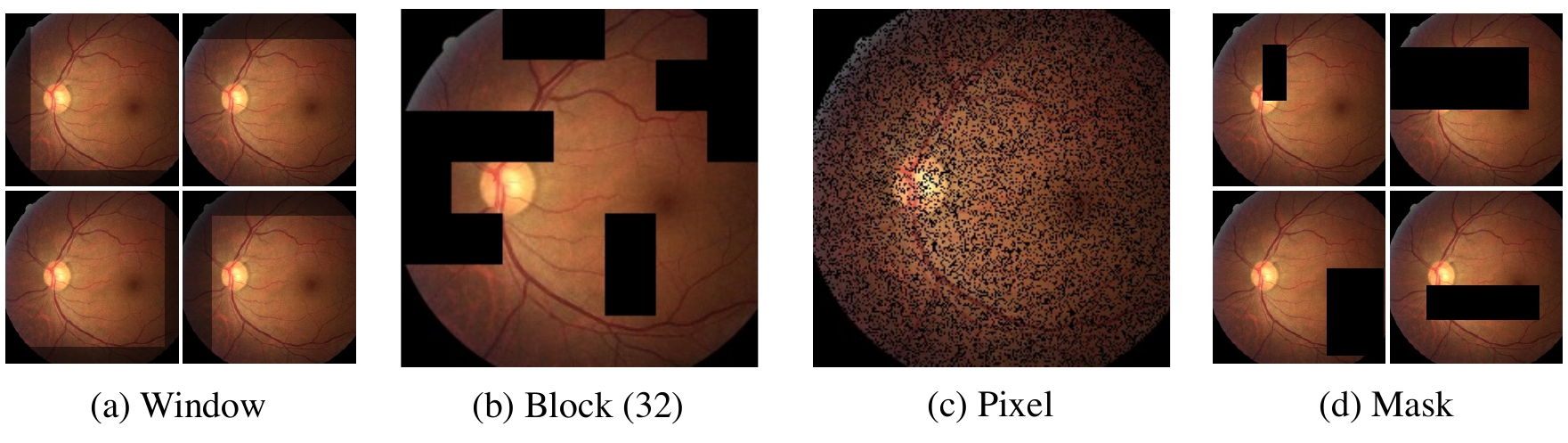}
    \vspace{-0.3cm}
    
    \caption{ Different Strategies for the computation of local statistics $\mu$ and $\sigma$. $\mu$ and $\sigma$ are computed on the bright area. They are conducted in the feature space actually, but here we demonstrate with the raw images.} 
    \label{fig:strategies} 
    \end{center}
    \vspace{-0.6cm}
\end{figure}

First of all, we removed the statistics mixing. Secondly, we discussed the choice of the statistics $\mu$ and $\sigma$ in \emph{WIN}. We employed the global mean $\mu$ or global variance $\sigma$ in the third and fourth row of Table \ref{tab:ablation}, respectively. Thirdly, we investigated the area contributed to $\mu$ and $\sigma$ (see Fig. \ref{fig:strategies}). The \emph{global} computes the mean and variance with all pixels, namely \emph{IN}. The \emph{Block} and \emph{Pixel} compute them with randomly selected blocks or pixels, respectively. The \emph{Mask} randomly erases a certain region and computes the $\mu$ and $\sigma$ with the remaining pixels. Finally, we perturbed the global statistics in another way, adding the speckle noise (namely the \emph{Speckle}) \cite{hendrycks2021many}. As shown in Table \ref{tab:ablation},  the following conclusions can be drawn: 1) Statistics mixing is beneficial. 2) Only employing one local statistic dramatically degenerates the OOD generalization. 3) \emph{Window} and \emph{Block} are the best practice for the local statistics computation. 4) Compared with \emph{IN}, perturbing the statistics with the speckle noise or other local statistics does not help the OOD generalization.

In \emph{WIN-WIN}, we removed the statistics mixing and consistency constraint (i.e., the first row of Table \ref{tab:ablation}), respectively. As shown in Table \ref{tab:ablation}, either of them helped the OOD generalization marginally. The best practice is to combine them together since two mechanisms can complement each other well. Furthermore, we imposed the consistency constraint on the features extracted from the penultimate layer using the NT-Xent loss \cite{gao2021simcse}. The gain of consistency constraint still existed but was inferior to the current design $\mathcal{L}_{JSD}$.

\begin{figure}[tb]
    \begin{center}
    \includegraphics[width=\linewidth]{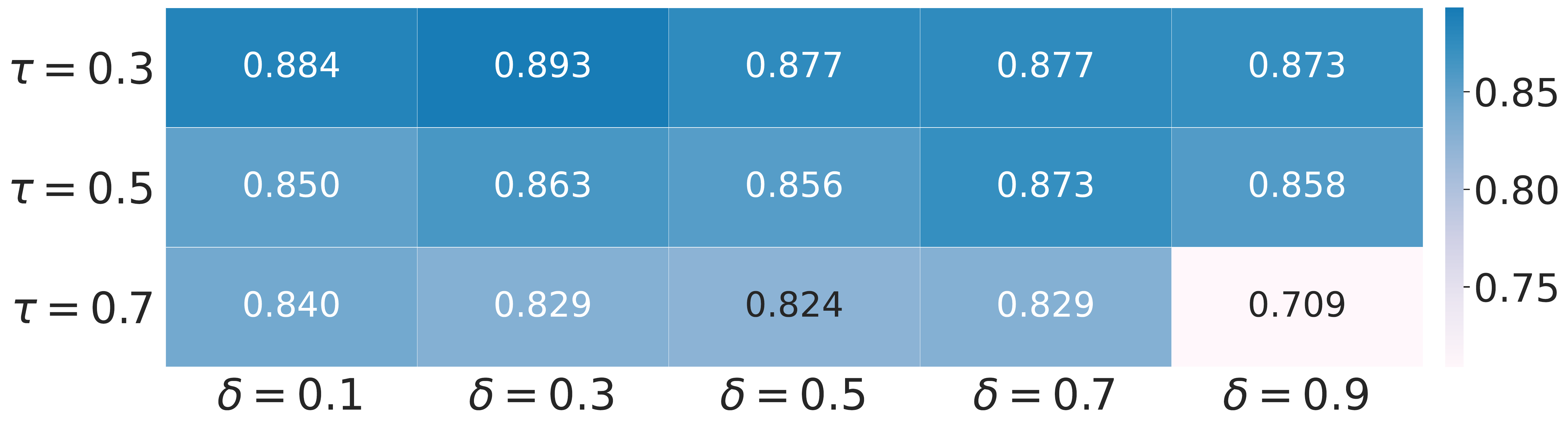}
    \vspace{-0.8cm}
    \caption{ Hyper-parameter sensitivity analysis for the $\delta$ and $\tau$.} 
    \label{fig:hyper-param} 
    \end{center}
    \vspace{-0.9cm}
\end{figure}

Fig. \ref{fig:hyper-param} shows the results of different hyper-parameters settings. The $m$-$cAUC$ was impacted by the window ratio threshold $\tau$ mainly, but was insensitive to $\delta$ (see Eq. \ref{eq:total}).  Choosing $\tau \in \{0.5, 0.7\}$ and $\delta \in \{0.1, 0.3, 0.5, 0.7, 1.0\}$ can gain a better result than \emph{IN}. 

\begin{figure*}[htb]
    \begin{center}
    \vspace{-0.1cm}
    \includegraphics[width=\linewidth]{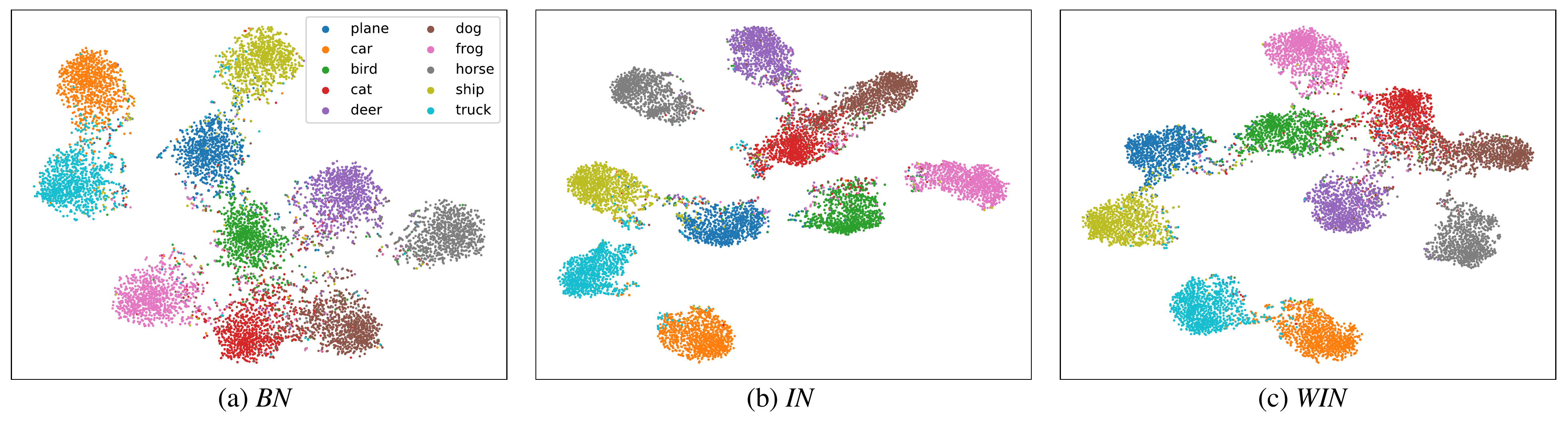}
    \vspace{-0.7cm}
    \caption{ T-SNE visualization of features on CIFAR-10-C. All plots are drawn with the features of in-distribution data (i.e., validation set).} 
    \label{fig:t-sne} 
    \end{center}
    \vspace{-0.6cm}
\end{figure*}

    
    

\begin{figure*}[htb]
    \begin{center}
    \includegraphics[width=\linewidth]{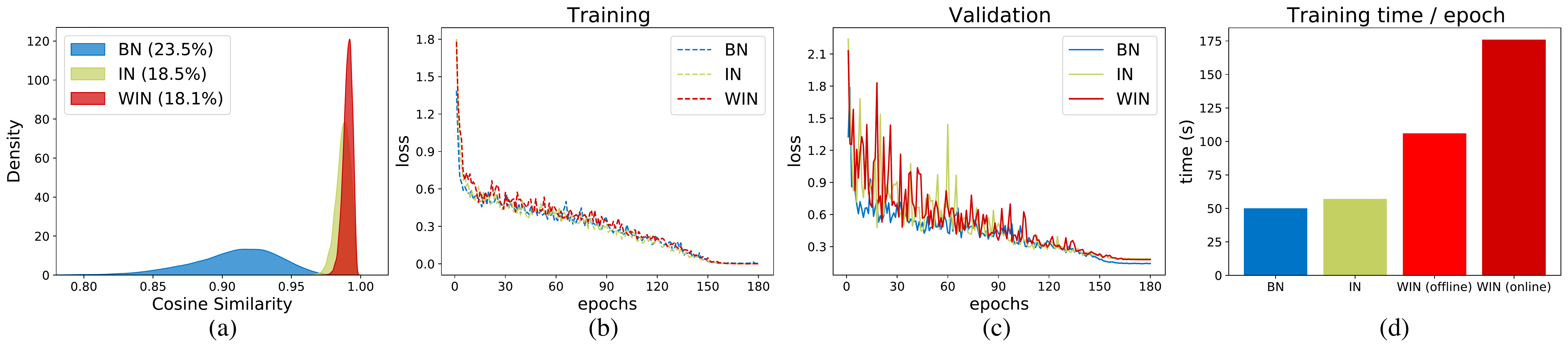}
    \vspace{-0.8cm}
    \caption{ (a) cosine similarity between the raw image and its corrupted versions (i.e., OOD data); (b) training loss curves on CIFAR-10-C; (c) validation loss curves on CIFAR-10-C; (d) time costs per epoch of ResNet-18 on CIFAR-10-C.} 
    \label{fig:training} 
    \end{center}
    \vspace{-0.9cm}
\end{figure*}

\subsection{Comparison with State-of-the-arts}
The comprehensive comparison with state-of-the-art methods is presented in Table \ref{tab:glaucoma}. All results are reported with the results of ResNet-50 on the glaucoma detection task.

We first compared our methods with several normalization methods. Among these methods, \emph{IN} \cite{DBLP:conf/icml/IoffeS15}, \emph{BN} \cite{DBLP:journals/corr/UlyanovVL16}, and \emph{GN} \cite{wu2018group} are the most popular normalization methods, LCN \cite{ortiz2020local}, SNR \cite{jin2021style}, BIN \cite{nam2018batch}, SN \cite{luo2019switchable}, IBN-a \cite{DBLP:conf/eccv/PanLST18}, and IBN-b \cite{DBLP:conf/eccv/PanLST18} are incremental studies for \emph{IN}, and RBN \cite{gao2021representative} and CNSN \cite{tang2021crossnorm} are recently proposed methods for improving the model generalization. Overall, these normalization methods did not present any obvious advantages for IND generalization and OOD generalization and our methods significantly outperformed them. It is worth noticing that LCN and CNSN are most relevant to our method. LCN normalizes every feature with the statistics of its neighbors. CNSN proposes the crossnorm which exchanges local statistics and global statistics between channels. Although local statistics are also used in LCN and CNSN, \emph{WIN} surpassed them considerably ($+0.104$ $m$-$cAUC$ for LCN and $+0.237$ $m$-$cAUC$ for CNSN) and this advantage was expanded by \emph{WIN-WIN} further.

On the other hand, our methods are also relevant to augmentation-based methods. The comparison with these methods also have been conducted using the ResNet-50 with \emph{BN}, including image-space augmentation (i.e., AutoAugment \cite{cubuk2019autoaugment}, RandAugment \cite{cubuk2020randaugment}, Patch Gaussian \cite{lopes2019improving}, randomErasing \cite{zhong2020random}, mixup \cite{DBLP:conf/iclr/ZhangCDL18}, CutMix \cite{yun2019cutmix}, AugMix \cite{DBLP:conf/iclr/HendrycksMCZGL20}, and PixMix \cite{hendrycks2022robustness}) and feature-space augmentation (i.e., manifold mixup \cite{verma2019manifold} and MixStyle \cite{zhou2021mixstyle}). As shown in Table \ref{tab:glaucoma}, these methods showed their superiority on IND data, but in terms of OOD generalization, their effects were negligible or even negative except for RandAugment and PixMix. Overall, our methods significantly outperformed these augmentation-based methods.

Furthermore, it should be noted that SNR, IBN-a, IBN-b, CNSN, Patch Gaussian, AugMix, PixMix, and MixStyle are state-of-the-art methods of domain generalization and adaptation or robustness which improve the OOD generalization with a single source domain. However, our methods significantly outperformed them.

\section{Discussions}

In practice, the training data is a small population of real data, which is smaller in medical images due to the expensive costs for data acquisition and the high diversity caused by operating procedures and imaging equipment. Thereby, the discrepancy between training data and test data from unseen deploy environments is prevailing exists and leads to performance dips. In this paper, we mainly focus on improving the OOD generalization on heterogeneous data which has the same labels but different appearances. A simple yet effective normalization technique \emph{WIN} is proposed to address this problem with free parameters and a negligible effect for IND data. \emph{WIN} conducts the feature-level augmentation by perturbing the normalizing statistics with stochastic window statistics. With this feature-level augmentation technique, we propose a novel self-distillation scheme \emph{WIN-WIN} to eliminate the train-test inconsistency and further improve the OOD generalization in classification tasks. \emph{WIN-WIN} is easily implemented by twice forward passes and a consistency constraint. Extensive experiments demonstrated our methods can significantly and generally boost the OOD generalization across different tasks spanning 24 datasets.

Here, we presented a comprehensive investigation of the properties of the \emph{WIN}, covering the impacts on IND data and OOD data, the model convergence, and the time costs.  To ensure reproducibility and provide insight for subsequent studies, we conducted the experiments on the widely-used OOD generalization benchmark CIFAR-10-C \cite{DBLP:conf/iclr/HendrycksD19}. Fig. \ref{fig:t-sne} and \ref{fig:training} (a) show the t-SNE visualization and the averaged cosine similarity between the raw image and its corrupted versions using the penultimate layer feature of ResNet-18, respectively. Although \emph{BN} reported the best result on IND data (see Table \ref{tab:cifar}), \emph{IN} and \emph{WIN} were better separated for different classes and showed a better invariance (\emph{WIN} is the best.). In addition, we analyzed the loss curves for \emph{BN}, \emph{IN}, and \emph{WIN} and their time costs. Fig. \ref{fig:training} (b) and (c) show that the model convergence rate of \emph{WIN} is not inferior to that of \emph{BN} and \emph{IN}. For the time costs, we introduced two schemes: \emph{WIN (offline)} and \emph{WIN (online)}. \emph{WIN (offline)} caches the window parameters, while \emph{WIN (online)} generates these on-the-fly. According to Fig. \ref{fig:training} (d), \emph{WIN (offline)} greatly reduced the time cost but was over $47.2 \%$ for \emph{BN} and  $46.3\%$ for \emph{IN}. In a word, despite the longer training time, \emph{WIN} can be a good alternative to existing normalization techniques, which enhances the feature representations and improves the OOD generalization effectively.

Despite its superior improvement in the OOD generalization, \emph{WIN} still has a limitation that the window sampling is inefficient for the inputs with a small size. Fortunately, it is no longer a limitation for existing CNNs. Since large input sizes can improve the model generalization \cite{tan2019efficientnet}, existing CNNs commonly use a large input size. As a variant of \emph{IN}, a reasonable conjecture is that the advantages of \emph{IN} are also retained in \emph{WIN}. In future work, we will apply \emph{WIN} to the pixel2pixel tasks (e.g., super-resolution and style transfer) that widely use the \emph{IN}. Furthermore, since its superiority for heterogeneous data, its use in federated learning is worth digging into thoroughly. 

\section{Conclusions}
In this paper, we propose a simple yet effective normalization technique, namely \emph{WIN}, which uses stochastic local statistics to boost the OOD generalization and not sacrifice the IND generalization. Based on \emph{WIN}, we propose the \emph{WIN-WIN} to further improve OOD generalization for classification tasks, which can be implemented with only a few lines of code. Our methods gracefully address the OOD generalization of heterogeneous data in real-world clinical practice. Extensive experiments on various tasks and datasets demonstrated their generality and superiority compared with the baselines and many state-of-the-art methods.

\bibliography{IEEEabrv, example}

\end{document}